\documentclass{article}
\usepackage{spconf,amsmath,graphicx}
\usepackage{booktabs}
\usepackage{makecell}
\usepackage{amsfonts}
\usepackage{multirow}
\usepackage{stfloats}
\usepackage{algorithm}
\usepackage{algorithmic}
\usepackage[colorlinks,
            linkcolor=blue,
            anchorcolor=blue,
            citecolor=black]{hyperref}
\usepackage[numbers, sort]{natbib}

\title{Category-Agnostic Pose Estimation for Point Clouds}
\name{Bowen Liu$^{1}$$^{3}$\qquad Wei Liu$^{1}$$^{3}$\qquad Siang Chen$^{2}$$^{3}$\qquad Pengwei Xie$^{2}$$^{3}$\qquad Guijin Wang$^{2}$$^{3}$}
\address{$^{1}$ Tsinghua Shenzhen International Graduate School\\
$^{2}$ Shanghai Artificial Intelligence Laboratory\\
$^{3}$ Tsinghua University, Department of Electronic Engineering}
\begin{document}
\topmargin=0mm
%
\maketitle
\begin{abstract}
The goal of object pose estimation is to visually determine the pose of a specific object in the RGB-D input. Unfortunately, when faced with new categories, both instance-based and category-based methods are unable to deal with unseen objects of unseen categories, which is a challenge for pose estimation. To address this issue, this paper proposes a method to introduce geometric features for pose estimation of point clouds without requiring category information. The method is based only on the patch feature of the point cloud, a geometric feature with rotation invariance. After training without category information, our method achieves as good results as other category-based methods. Our method successfully achieved pose annotation of no category information instances on the CAMERA25 dataset and ModelNet40 dataset.\par
\end{abstract}
\section{Introduction}
\label{sec:intro}
\begin{figure}[t]
\centering
\includegraphics[scale=0.29]{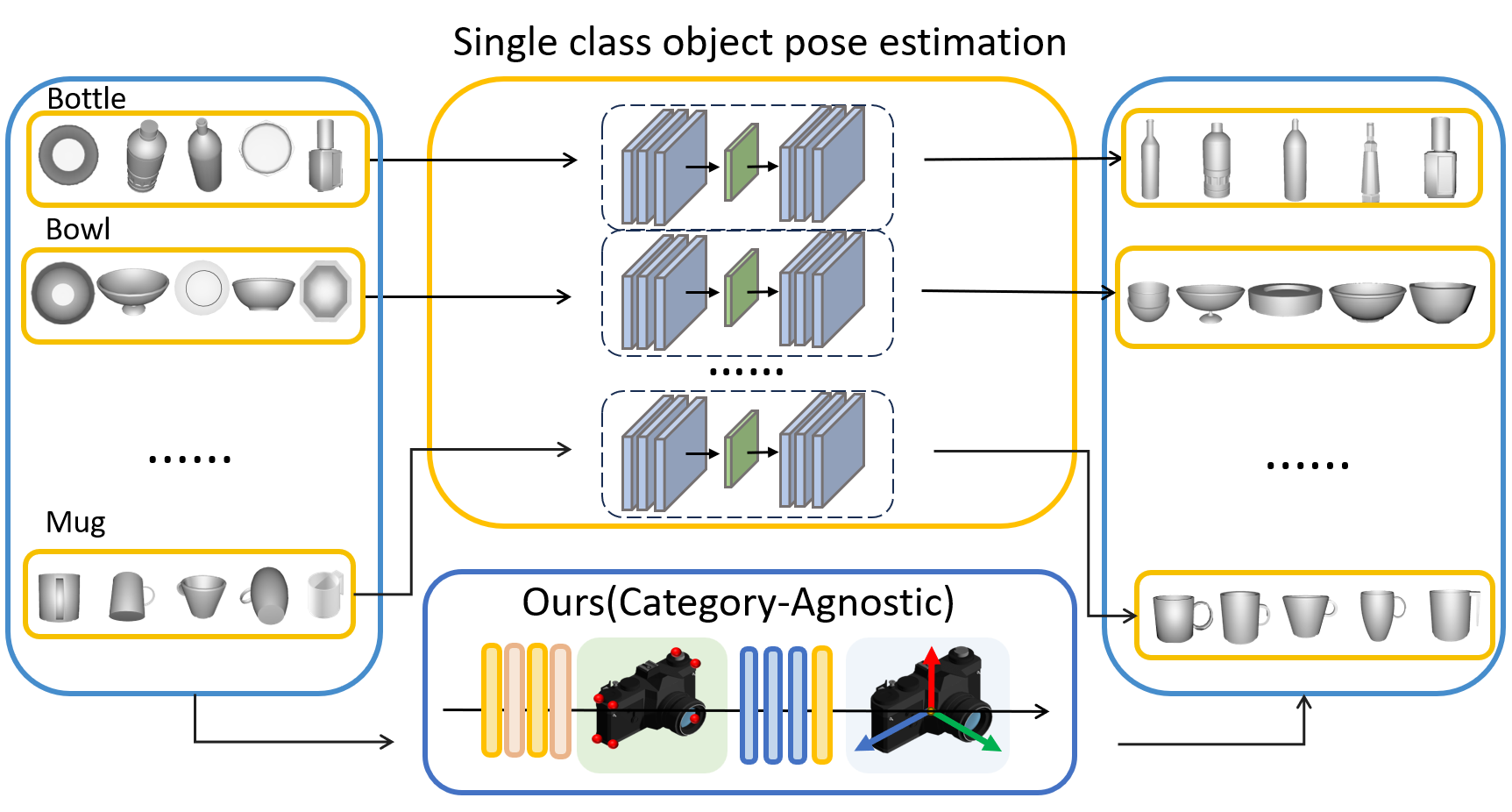}
\caption{Most methods of pose estimation based on a single category are complicated and difficult to generalize to other categories. To address this issue, we propose a method based on geometric features for category-agnostic pose estimation.}
\label{issue}
\end{figure}

Object pose estimation is a crucial intermediate step in fields such as augmented reality, autonomous driving, and robot grasping\cite{du2021vision}. Pose estimation evaluates the orientation of objects and utilizes this information for downstream tasks. However, due to the need for pose annotation during training\cite{huang2013fine}, inconsistencies arise between different datasets, which lead to the failure of downstream tasks\cite{sedaghat2016orientation}. Moreover, for example, in Fig.\ref{issue}, there has been limited focus on training models that can generalize across categories or datasets, making it challenging to conduct comprehensive experiments in pose estimation tasks.\par
Most of the dataset models for pose estimation work are derived from ShapeNet \cite{chang2015shapenet}, which relies on manual filtering and component semantic alignment for intra-category pose alignment, making it unsuitable for working on unlabeled or uncategorized models. To address the alignment reference framework for category-level objects, EPN \cite{chen2021equivariant} and SE(3) \cite{li2021leveraging} perform pose alignment within the same category. However, these methods using equivariant networks can only estimate pose based on the unified geometric features of objects of the same category, and cannot provide reasonable pose estimation for new categories.\par
Therefore, to address the obstacles of missing category information to pose estimation work, we start from the geometric features inherent in desktop-level objects. This approach for estimate poses is intuitive, interpretable, and stable category-agnostic. This feature can predict object poses after being fed to our pose estimation networks. Compared to other schemes that train networks separately for pose estimation based on objects of the same category, our scheme can achieve results that are comparable to the above schemes by only utilizing geometric features. The experimental results on the CAMERA25\cite{wang2019normalized} dataset and ModelNet40\cite{wu20153d} dataset have validated the efficiency and stability of this method. Experimental results have shown that the features significantly enhance the network's ability to predict the orientation of objects. To the best of our knowledge, this paper is the first to attempt to predict the pose of desktop-level objects based only on point clouds, without category information.\par
In summary, our contributions can be outlined as follows:\par
1. We propose an end-to-end pipeline to automate the point cloud pose estimation without category information.\par
2. We propose a geometric feature with rotation invariance called patch, and design a semi-automated method to achieve the annotation of the training dataset.\par
3. We conducted experiments on different datasets to demonstrate the feasibility of a pose estimation pipeline based on patch features without categorical information.
\section{RELATED WORK}
\label{sec:relatedwork}
The pose estimation based on the RGB-D input can be regarded as the part-to-whole registration problem from the single view point cloud obtained from the RGB-D image to the existing complete object point cloud. This type of problem usually requires a 3D model of the object or a point cloud as an estimation template. Common methods are either based on regression \cite {rad2017bb8, kehl2017ssd} to find the internal relationship between pose and image, or to determine the final pose by matching it with a standard template of predefined poses \cite{xiang2017posecnn, park2020latentfusion,xiang2017posecnn}. To solve the occlusion problem, there are also some works using voting estimation \cite{wang20206, he2020pvn3d} to match the image pixel by pixel to 3D coordinates, and finally using Perspective-n-Point (PnP) \cite{fischler1981random} to calculate the pose.\par
With the maturity of 6D pose estimation for instance-level objects, some methods of 6D pose estimation for category-level objects have emerged. These methods use the same category object for learning, and can estimate the pose of unseen objects of that category, which require that objects of the same category in the training data have a highly consistent pose definition.  Traditional single-target estimation datasets based on manual annotation, such as LineMOD \cite{hinterstoisser2011multimodal} and YCB-Video \cite{calli2015ycb}, do not strictly define the target pose, and only use simple geometric features or habits as the standard rotation estimation template. As a result, the pose definitions of different objects in the same category are difficult to be completely consistent, and there are differences among different datasets. The ShapeNet model dataset first addresses this problem and attempts to use semantic information and manual methods, as well as feature-based semi-automatic methods for in-category object alignment.\par
Recently, some works using invariance or equivariant of point cloud\cite{poulenard2019effective, fuchs2020se3transformers, li2021leveraging} to estimate the pose automatically, try to break through the limitation of the need for intra-category object alignment in dataset. Different with the idea of focusing on pose estimation of intra-class objects, we hope to achieve a wider range of pose estimation through geometric features.\par
To sum up, the category-level pose estimation methods need to classify the data using the category information first, and then perform the pose estimation separately, which makes it difficult to deal with a large number of unseen objects from new categories in the real scene.   Moreover, these methods require the use of semantic information or manual methods to align the poses of objects within a category, which is costly and inefficient. To address this issue, we design patches based on the geometric characteristics of objects with rotation invariance, which successfully avoids the dependence on category information.\par
\section{METHOD}
\label{sec:typestyle}
We propose a pose estimation method based on the geometric features of the point cloud, which focuses on the rotation invariance feature called patch (see section \ref{annotationProcess} for details).  It can achieve pose estimation of the target without category information and only requires semi-automatic annotation of the dataset before training the network. Section \ref{deeplearningmethod} provides an overall introduction to the pipeline, including patch prediction and pose estimation.  Section \ref{annotationProcess} shows how we obtain patches in the dataset as training inputs.
\subsection{End-to-end pose estimation}\label{deeplearningmethod}
\begin{figure*}[ht]
\centering
\includegraphics[scale=0.35]{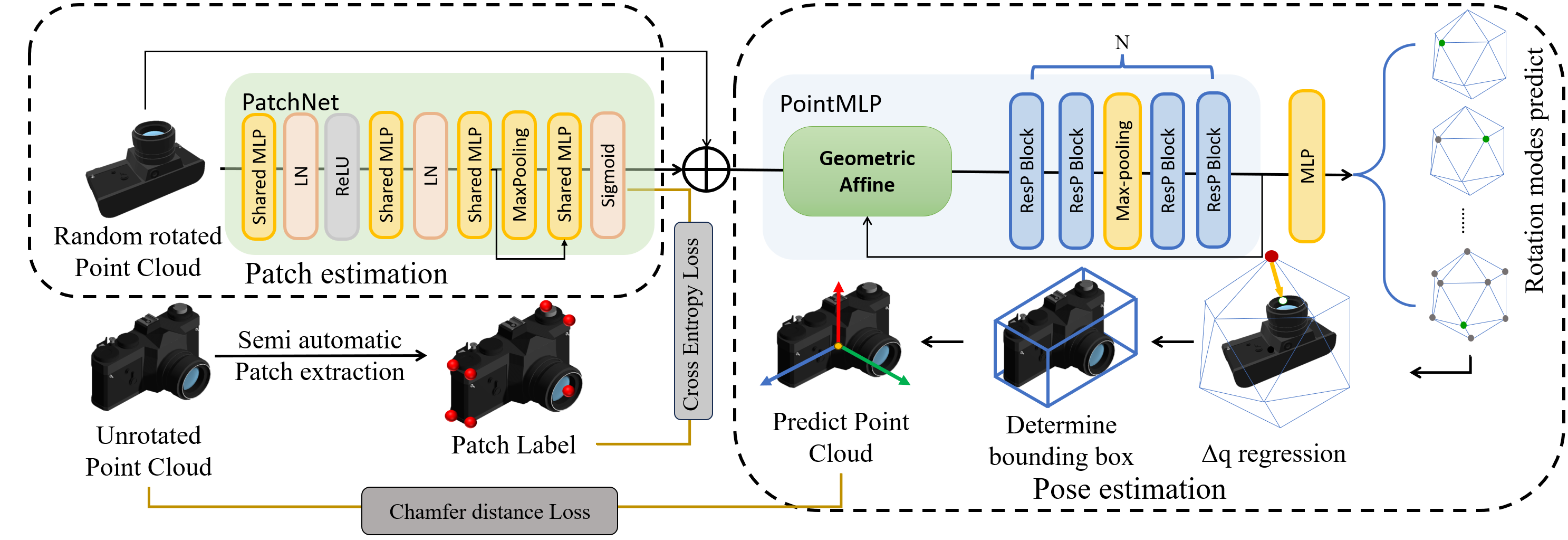}
\caption{Point cloud pose estimation pipeline (to enhance readability, we employ colored models instead of point clouds).: The randomly rotated point cloud $\mathbf Y$ is used as the input of the network. PatchNet is responsible for predicting the patch of the point cloud to generate geometric features, which are combined with the global point cloud and input into the network with PointMLP as the backbone. Finally, output $\Delta \mathbf q$ to predict and compensate 60 rotation modes of the icosahedron to get the corrected pose.
The network measures the Loss between point clouds through chamfer distance.}
\label{pipline_all}
\end{figure*}
Fig.\ref{pipline_all}  shows the structure of our pipeline, including patch estimation part and pose estimation part.
In the training process, the patch obtained by the semi-automatic method (see Section \ref{annotationProcess} for details) is used as the ground truth of the PatchNet.
In the inference stage, PatchNet takes the point cloud as input and automatically completes the extraction of patches.
In the pose estimation part, the point cloud and predicted patch are input to the PointMLP \cite{ma2022rethinking} for estimation, and the space is divided into 60 predicted basic poses to provide initial direction estimation, avoiding over-parameterization.
(due to the nonlinearity of Euler angles in rotational space and the computational intensity of continuous space calculation \cite {huynh2009metrics}). In the final loss design, $\Delta \mathbf q$ is introduced as a small rotation for pose regression.

\subsubsection{Patch Estimation}
Before feeding the point cloud to PatchNet, the CAD model mesh is converted to a fixed number of point clouds ($N$=1024) using FPS sampling, and the range of coordinate extremes is selected as the unit size to normalize its size.\par
In the patch estimation part, the patch network extracts patch features of the point cloud and continuously focuses on the patch parts at a global scale of the point cloud through multi-layer shared parameter MLP. This structure consists of two layers of MLP with shared parameters, so it does not overly focus on the local details of the point cloud. It only predicts patch spatial features at the global scale to prevent overfitting on specific geometric structures. \par
Finally, the patch points preprocessed in Section \ref{annotationProcess} are used as the ground truth, and the predicted patch probability of each point is evaluated using the cross-entropy loss, which is a part of the total loss used for end-to-end training.
The network predicts and outputs a fixed number of patch points
The downstream PointMLP will extract the multi-scale local features of the point cloud, and since our approach is based on the geometric features of the point cloud, it is natural to expect the patch features to guide the network.\par

\subsubsection{Pose Estimation}\label{poseEstimation}
Inspired by SE(3)\cite{li2021leveraging} method, the pose estimation section first classifies the rotation mode of the target, and then performs local error regression.
Since a target object can possess multiple rotation modes in the icosahedron rotation group $\mathbf G$, determining the definitive standard pose is challenging. Therefore, in the pose estimation process, it is crucial to estimate the probability distribution of $\mathbf{ g \in G}$ while aligning it with the actual spatial orientation. To achieve this, we discretize the space rotation angles and reduce the search space by selecting 60 rotation modes, which correspond to the Regular Icosahedron and its corresponding quaternion $\mathbf q$. However, introducing the discretized quaternion may result in a small error, $\Delta \mathbf q$, in the final accuracy. It can be demonstrated that the error between nearest-neighboring elements from icosahedron rotational group $\mathbf G$ is $\pi/5$. Consequently, the compensation  $\Delta \mathbf q$, obtained in the final predicted value must be smaller than $\pi/5$. 
To avoid over-parameterization of $\Delta \mathbf q$, a lower limit of $q_w$ is set to ensure that $q_w > \cos (\pi/10)$.
Taking these considerations into account, the following constraints are applied:
\begin{equation}
||\Delta \mathbf q|| = \cos(\pi/10)+(1-\cos(\pi/10)) \cdot sigmoid(||\Delta \mathbf q||),
\end{equation}
this formula applies to the unit Quaternion, and the above two points are realized through the Sigmoid function.
\subsubsection{Loss Function}
Between the given point cloud $\mathbf X$ and its rotated point cloud $\mathbf Y$, there exists a rotated pose $\mathbf{P=(q,t)}$,
where $\mathbf {q,t}$ represent the Quaternion of the pose 
 and the coordinate difference between them. Consequently, the relationship between $\mathbf X$ and $\mathbf Y$ can be expressed as: $\textbf{Y}=\textbf{qXq}^{-1}+\textbf{t}$. 
 In this case, the mean coordinates of the point clouds are used as rotation centers, since only rotational correspondences are considered.
 Theoretically, this center corresponds to the center of the object when employing farthest point sampling and maintaining rotation invariant.\par
Based on the above, we consider the minimum Euclidean distance loss of $N$ points under the rotation index.
Therefore, the chamfer distance\cite{butt1998optimum} is employed:
\begin{equation}
    d(\mathbf{X,Y})=\frac{1}{N_\mathbf X} \underset{{\boldsymbol{x} \in {\mathbf X}}}{\sum} \underset{\boldsymbol{y} \in {\mathbf Y}}{\min}||\boldsymbol{x}-\boldsymbol{y}||_2^2+\frac{1}{N_\mathbf Y} \underset{{\boldsymbol{y} \in {\mathbf Y}}}{\sum} \underset{\boldsymbol{x} \in {\mathbf X}}{\min}||\boldsymbol{y}-\boldsymbol{x}||_2^2,
\end{equation}
where $\boldsymbol{x}$,$\boldsymbol{y}$ are points in point cloud $\mathbf{X}$,$\mathbf{Y}$.\par
Additionally, regarding $\Delta \mathbf q$ in the rotation mode, since the unit quaternion must correspond to the rotation mode, we normalize the final output quaternion. We adopt the square distance metric, denoted as $L_q=(|| \Delta \mathbf q||_2-1) ^ 2$, for this purpose. In summary, the final loss is expressed as 
\begin{equation}
 L=L_{pose\_rec}+\lambda_1 L_q+ \lambda_2 L_{patch},
\end{equation}
where $\lambda_1$, $\lambda_2$ represents the regularization weight, $L_{pose\_rec}$ is the chamfer distance between the point cloud and the origin cloud under the final estimated rotational pose, and $L_{patch}$ is Cross Entropy Loss between patch labels and PatchNet prediction.
\subsubsection{Symmetry metric}\label{Measurement of object}
Due to the utilization of category-irrelevant point clouds for pose estimation, the metrics employed in the loss design process solely relate to the point-to-point distance between point clouds. Unfortunately, this metric is influenced by the actual geometric shape of the object, causing instances in the same class to be interfered with by highly similar symmetrical structures.\par
To address this concern, we extend the ground truth to include other symmetric axes in the spatial coordinate system. As a result, estimation results of the network encompass the symmetric directions, including vertical or opposite rotations, within the statistics. This approach ensures that the network focuses on accurately estimating the pose of the object, rather than rigidly fitting the ground truth.\par
In the end, we will use a unified projection rule to determine the orientation, ensuring consistency in orientation among objects of the same category. The specific method involves projecting each point $\mathbf p_i$ in the point cloud in the current orientation direction, determining the distance $D_i$ of each point to the projection plane at the origin $O$, where $i \in \left[1, N\right]$. The sum is then calculated along each axis, and the three resulting positive or negative values determine whether the final orientation is inverted. This method is insensitive to noise, focusing more on the influence of the distant parts on the orientation.
\subsection{Patch Feature of Dataset}\label{annotationProcess}
We preprocess the dataset to obtain the patch features required for training. The following two sections respectively introduce the annotation methods and discuss the rotation invariance involved in this process. It is worth noting that the annotation requirement for patches only occurs during the network training phase, and is not required during the inference phase.
\subsubsection{Annotation of patch}
\begin{figure}[!ht]
\centering
\includegraphics[scale=0.3]{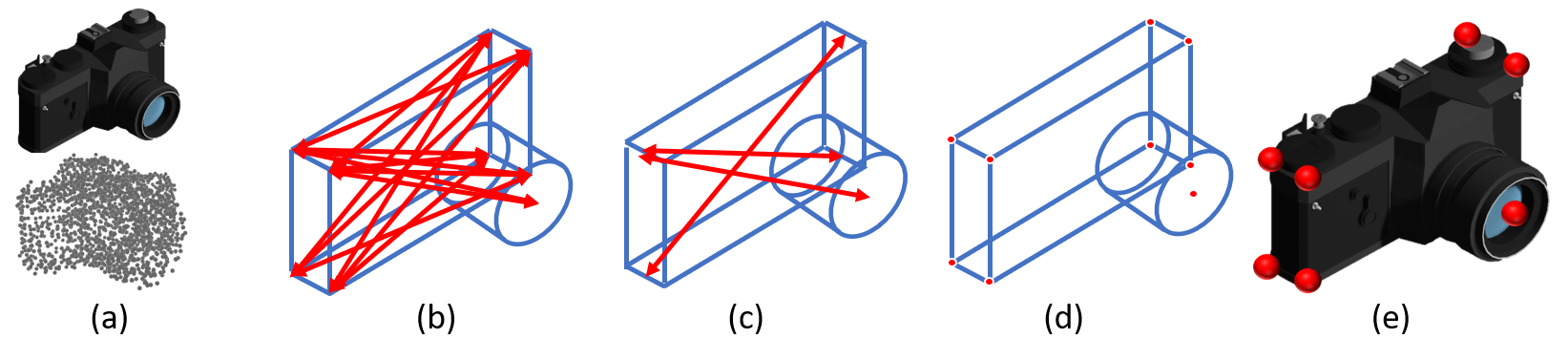}
\caption{The semi-automatic patch annotation process, where semi-automatic refers to some parameters being manually selected, may cause some differences in the results between different objects}
\label{patch_extraction}
\end{figure}
The entire processing flow is briefly shown in Fig.\ref{patch_extraction}, and the details are described as follows:\par
\begin{algorithm}
    \renewcommand{\algorithmicrequire}{\textbf{Require:}}
    \renewcommand{\algorithmicensure}{\textbf{Output:}}
    \caption{The semi-automatic patch annotation process}
    \begin{algorithmic}[1]
        \REQUIRE 
        $mesh\gets$ the CAD model, \par
        $N\gets$ the points number, \par 
        $th\gets$ the threshold, \par
        $M\gets$ the maximum number of feature vectors. \par
        \STATE $P=\mathtt{FPS}(mesh,N)$, get the point cloud by FPS,
        \STATE $\mathcal V =\mathtt{VectorInPC}(P)$, pairwise vectors formed from $P$,
        \STATE $\mathcal V \gets$ Sorted by $\mathcal ||\mathcal V||$ in descending order,
        \STATE $\mathcal V_M \gets$ The first $M$ elements of $\mathcal V$,
        \STATE $V_d =\mathtt{CosineDistance}(\mathcal V_M)$, cosine distance of $\mathcal V_M$,
        \STATE $\mathcal V_k=\mathtt{Cluster}(\mathcal V_M,V_d, th)$, cluster $\mathcal V_M$ by $\mathcal V_d$ and $th$,
        \STATE $\mathcal{P}_r=\mathtt{Pair}(\mathcal V_k)$, get endpoints from $\mathcal V_k$,
        \STATE $Patch=\mathtt{Query Ball Point}(\mathcal P_r$)
        \ENSURE $Patch$
    \end{algorithmic}
\end{algorithm}
1. Sampling the model mesh (Fig.\ref{patch_extraction}(a)) with a fixed number of $N$ points at the farthest distances. 
The farthest point sampling is adopted to ensure the spatial uniformity of the point distribution.
Typically, $N$ is chosen as 1024.\par
2. Setting a maximum limit of $M$ for the number of feature vectors and selecting the top $M$ feature vectors from the farthest distances. 
Here, the first M feature vectors are selected from the N farthest points (Fig.\ref{patch_extraction}(b)).
Generally, their orientations are defined positively along the z-axis in the initial coordinate system.\par
3. Filtering these $M$ feature vectors by clustering them based on the cosine similarity angle threshold $th$. The mean vectors within each cluster are chosen as the final direction vectors $\mathcal V_k$ with $k$ classes (Fig.\ref{patch_extraction}(c)). The number of $k$ typically ranges from 1 to $M$, and it is related to the value of $th$. For instance, when using 10° as the threshold, there are usually around 4 different orientations of $\mathcal V$ selected.\par
4. In an ideal scenario, the above steps would yield a corner based on object symmetry (Fig.\ref{patch_extraction}(d)). However, due to point cloud omissions or noise, some angular deviations along the axis are inevitably present. Therefore, choosing to include points near the feature points (Fig.\ref{patch_extraction}(e)) along these orientations acts as a blur, statistically mitigating the impact of individual angular errors. Finally, the feature points and the points around them form the patch of the point cloud.
\subsubsection{Rotation Invariant of Patch}\label{featureDesign}
In this section, we introduce the implementation of the above methods in rotation-invariance.
\begin{itemize}
    \item \textbf{Inference ability on the axis:} In desktop-level scenes, geometric features typically exhibit invariance along their direction or orthogonal directions. For instance, containers have a top-to-bottom direction, considering their vertical orientation when placed on a tabletop, and the perpendicular direction as the annotated orientation.
    \item \textbf{Edge convex region of objects:} The local geometric features of the object are the important basis for pose estimation. These Convex regions are the reference for identifying different directions and determining the orientation of the object, and they mainly appear at the distal end of the object. For example, the handle of a mug is an important feature that distinguishes it from a rotation-symmetric object, and these peripheral geometric features should be emphasized.
    \item \textbf{Robustness in point cloud:} Discussion about robustness in the face of noise first appeared in PointNet\cite{qi2017pointnet}. Sparse sets learned by networks demonstrate that an object only needs a skeleton to ensure the robustness and resilience of the point cloud feature model. The farthest point on the axis of the object illustrates this robustness. Only local erosion and missing features at the farthest point can affect them, but multiple selections at the farthest point and normal vectors can mitigate the impact of such losses.
\end{itemize}\par
The design process of the patch fully reflects the rotation invariant features mentioned above.
\section{EXPERIMENT}
\label{sec:EXPERIMENT}
\subsection{Dataset and Preprocess}
The models in two datasets in our work are primarily from ShapeNet, which offers various categories for model selection. Several popular models were chosen based on common dataset categories. For testing, we used popular objects such as bottles, bowls, cans, boxes, cameras, and mugs in the simulation scene of the CAMERA25 dataset. For comparison with several networks\cite{chen2021equivariant,thomas2019kpconv} in the SE(3) task, we selected cars, chairs, and sofas in ModelNet40. 
These datasets are randomly rotated using Euler angles and Gaussian noise with $\sigma=0.1$ is introduced to ensure robustness during training and testing.
The original training/testing partitioning of the CAMERA25 and ModelNet40 datasets was followed.\par
To verify the generalization ability of the network, we selected 3 unseen categories: monitor, phone, and laptop for testing. These categories were also from the CAMERA25 dataset, but they had rarely been used in previous work, with a total of 1203 instances. Importantly, these categories were not included in the training/validation set, ensuring the novelty of the features.
\subsection{Patch Annotation experiment and result}
\begin{figure}[ht]
\centering
\includegraphics[scale=0.45]{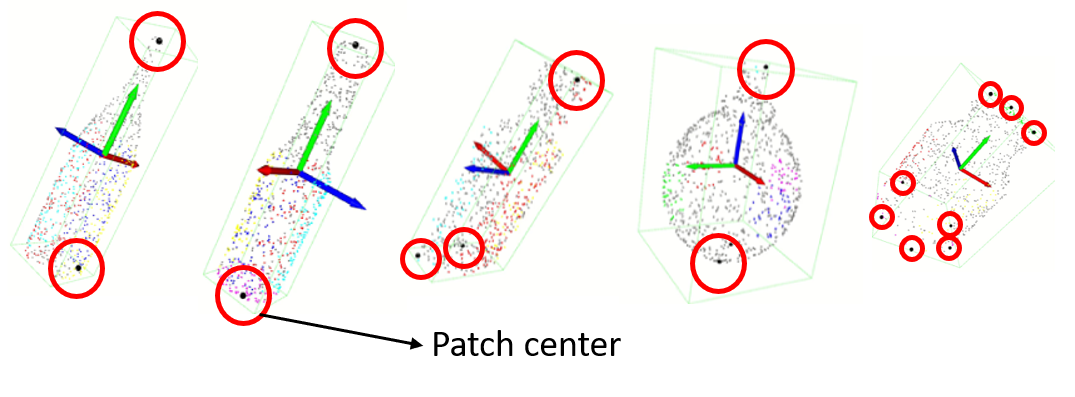}
\caption{The visualization of semi-automatic patch annotation process. To improve readability, we use the patch center to represent the patch area. The patch consistently appears at the top and bottom of the bottle category. This result was obtained under the conditions of $N=1024$, $M=20,$ and $th=10$.}
\label{patchexp}
\end{figure}

Random noise with $\sigma=0.1$ is added to the point cloud and rotated randomly, then the order of the point clouds is shuffled.
Repeat 10 times on each object in the CAMERA25 dataset. Ultimately, 960$\times$10 different point clouds were generated as samples, and the stability of the patch was verified by using the same parameters on each category. We use the position of the patch center as a criterion for determining stability. Result show 850 objects of all 960 objects maintaining the same patch center at least 8 times on each of 10 rotations. For the results of objects of the same category, the position of patch distribution is stable. Fig.\ref{patchexp} shows the stable patch we obtained on the bottle category, the center of each patch represented by black points. 
\subsection{Pose estimation experiment}
The experiments on the CAMERA25\cite{wang2019normalized} dataset and ModelNet40 dataset have demonstrated the possibility of patch features for estimating the pose of object point clouds from geometric structures without category information.\par
\begin{table*}[ht]
    \centering
    \scalebox{0.88}{
    \begin{tabular}
    {lccccc}
    \toprule
        \textbf{Mean(°)$\downarrow$ / Med.(°) $\downarrow$ / 5° mAP$\uparrow$ }& Airplane & Bottle & Car & Chair & Sofa  \\ \midrule
        EPN(category supervised)\cite{chen2021equivariant}  & 3.35/\textbf{1.12}/\textbf{0.96}&9.48/\textbf{1.85}/\textbf{0.95}&8.56/3.87/0.68&4.76/\textbf{1.56}/\textbf{0.97}&49.72/43.19/0.03   \\
        KPConv(category supervised)\cite{thomas2019kpconv}  &14.85/10.78/0.12&38.16/19.26/0.06&20.39/12.01/0.06  & 128.62/134.40/0.00&\textbf{1.33}/1.36/\textbf{1.00}   \\
        ICP + template shape & 8.11/1.22/0.90&22.76/2.94/0.69&88.92/96.28/0.10&39.00/9.69/0.28&76.11/54.42/0.04   \\
        SE(3)(category self-supervised)\cite{li2021leveraging} & 23.09/1.66/0.87  &17.24/2.13/0.89  &7.05/4.55/0.56&8.87/3.22/0.68&4.42/\textbf{0.83}/0.98   \\
        Ours(w/o category) & \textbf{2.56}/2.03/0.85 & \textbf{2.15}/2.09/0.86 & \textbf{2.30}/\textbf{2.06}/\textbf{0.87}&\textbf{3.65}/2.69/0.86&4.56/3.77/0.83  \\ 
        \bottomrule
    \end{tabular}
    }
    \caption{Pose estimation results for 5 categories in ModelNet40 Dataset. We report mean and median rotation errors and 5°mAP. The best performance is \textbf{Bold}.} 
    \label{Modelnet40Table}
\end{table*}
\begin{table}[ht]
    \centering
    \scalebox{0.68}{
    \begin{tabular}{l*{8}{c}}
    \hline
        \multirow{2}*{Category} & \multicolumn{5}{c}{Seen Category} & \multicolumn{3}{c}{Novel Category} \\ \cmidrule(lr){2-6} \cmidrule(lr){7-9}
        & Bottle & Bowl & Camera & Can & Mug & Monitor & Phone & Laptop  \\ \midrule
        Count & 235 & 140 & 75 & 55 & 150 & 453 & 359 & 391  \\
        Mean(°)$\downarrow$ & 2.16 & 2.23 & 2.78 & 2.32 & 2.88 & 2.73 & 2.57 & 2.45  \\ 
        Mid(°)$\downarrow$ & 2.11 & 2.03 & 2.32 & 2.26 & 2.58 & 2.43 & 2.23 & 2.20  \\
        5° mAP $\uparrow$ & 0.87 & 0.89 & 0.84 & 0.88 & 0.84 & 0.85 & 0.88 & 0.88  \\ \bottomrule
    \end{tabular}
    }
    \caption{Pose estimation in CAMERA25 Dataset. We report mean and median rotation errors and 5°mAP.} 
    \label{NOCSCategoryTable}
\end{table}
In this experiment, we performed pose alignment on point clouds of randomly rotating objects from the CAMERA25 dataset and ModelNet40 dataset, focusing on five widely recognized object categories without class information. The results are presented in Table \ref{NOCSCategoryTable}. The results indicate that methods that do not require category information still have good generalization performance on new categories, and the performance on the three new categories we use is encouraging. It is worth noting that since the laptop is a hinged object, its geometric form is only noteworthy in the direction perpendicular to the panel and the hinge axis. Therefore, we have modified the pose standard according to the above rules.\par
\begin{figure}[ht]
\centering
\includegraphics[scale=0.25]{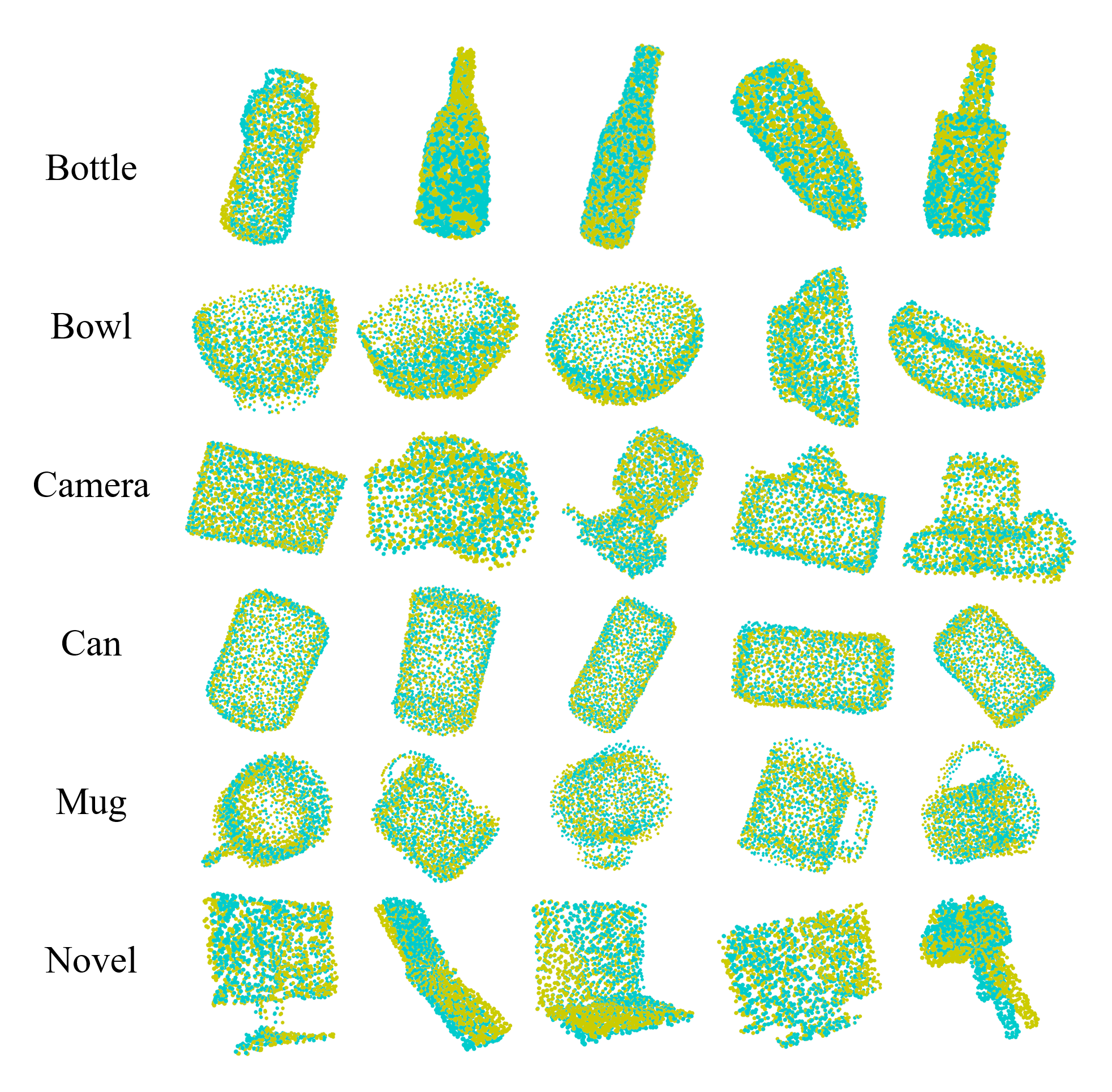}
\vspace{-0.1cm} 
\caption{The visualization of several categories of results in the CAMERA25 dataset, shows that in most cases, the results of pose estimation are satisfactory. On categories with significant geometric differences within, such as cameras, is challenging.}
\label{category_visual}
\end{figure}
Compared with EPN, KPCovn, and SE(3), our method integrates data from the ModelNet40 dataset and produces prediction results for all categories, as shown in Table \ref {Modelnet40Table}. EPN, as a network specifically studying the rotation of single-class objects, has achieved leading results in most categories, but it has a disadvantage in dealing with objects with blurred local details. The SE(3) network serves as a reference for pose prediction in this scheme, and its performance is similar to that of EPN in most categories. However, it still needs to be trained on objects in each category, and cannot achieve prediction outside of that category. Therefore, although our method is not state-of-the-art, it provides a good idea for pose prediction based on geometric features.\par

\begin{table}[ht]
    \centering
    \scalebox{0.8}{
    \begin{tabular}
    {lc}
    \toprule
        \textbf{Mean(°)$\downarrow$ / Med.(°) $\downarrow$ / 5° mAP$\uparrow$ }& Avrage on all categories \\ \midrule
        Pose estimation(PointMLP only) & 24.59/23.35/0.01 \\
        Ground Truth Patch + Pose estimation &  2.87/2.45/0.87 \\
        Patch estimation + Pose estimation & 3.04/2.53/0.85 \\ 
        \bottomrule
    \end{tabular}
    }
    \caption{Comparison of results in three approaches. We use Modelnet40 dataset, and the result is the average value across 5 categories} 
    \label{Ablation}
\end{table}
Fig.\ref{category_visual} shows our visualization in the CAMERA25 dataset. 
Define an unrotated point cloud as $\mathbf{P}_o$, the random rotation matrix as $\mathbf{R}$, and the estimated rotation matrix as $\mathbf{P}$. The point cloud $\mathbf{R \cdot P}_o$ obtained by random rotation and the point cloud $\mathbf{P\cdot P}_o$ obtained by predicted pose are visualized. The results show that even if only patch features are introduced, the result of target pose estimation can still be comparable to that of the dedicated estimation network using only a single category. In the symmetric category, our method has obtained satisfactory results for the prediction of key axes. For classes with special structures, such as the mug class, the patch gives weight to the handle, which makes our network perform well in that direction. Which is a challenge for other networks that align point clouds. At the same time, we also noticed that due to the significant differences in geometric shapes within the class, objects in the camera category perform well on their main axes and can still provide relatively stable results even in cases of drastic changes in shape, indicating that the network's application of geometric features in pose estimation is successful. The object in the last row is a category that has never been seen during training, indicating that the generalization ability of the pose estimation method based on geometric features is very promising.
\subsection{Ablation experiment}
Our ablation experiment design consists of three parts: the first part is to delete the patch and directly input the point cloud into PointMLP for pose prediction to verify the impact of the patch on pose estimation; The second part replaces the original prediction patch network with the ground truth of the patch to verify the effectiveness of patch estimation. The third part presents the entire pipeline we propose. The results are shown in Table \ref{Ablation}, and the ablation experiment strongly proves the effectiveness of the patch part.
\section{CONCLUSION}
\label{sec:CONCLUSION}
We attempted to use only geometric features for pose estimation of the target point cloud, and designed a complete automatic pose estimation scheme based on this feature. The method can estimate the pose of the point cloud without requiring category information. Satisfactory results were achieved on the ShapeNet-based model. We also hope that future work can further obtain this feature on incomplete and local point clouds and achieve pose estimation tasks.
\bibliographystyle{ieeetr}
\bibliography{IEEEabrv, refs}

\begin{thebibliography}{10}

\bibitem{du2021vision}
G.~Du, K.~Wang, S.~Lian, and K.~Zhao, ``Vision-based robotic grasping from
  object localization, object pose estimation to grasp estimation for parallel
  grippers: a review,'' {\em Artificial Intelligence Review}, vol.~54, no.~3,
  pp.~1677--1734, 2021.

\bibitem{huang2013fine}
Q.-X. Huang, H.~Su, and L.~Guibas, ``Fine-grained semi-supervised labeling of
  large shape collections,'' {\em ACM Transactions on Graphics (TOG)}, vol.~32,
  no.~6, pp.~1--10, 2013.

\bibitem{sedaghat2016orientation}
N.~Sedaghat, M.~Zolfaghari, E.~Amiri, and T.~Brox, ``Orientation-boosted voxel
  nets for 3d object recognition,'' {\em arXiv preprint arXiv:1604.03351},
  2016.

\bibitem{chang2015shapenet}
A.~X. Chang, T.~Funkhouser, L.~Guibas, P.~Hanrahan, Q.~Huang, Z.~Li,
  S.~Savarese, M.~Savva, S.~Song, H.~Su, {\em et~al.}, ``Shapenet: An
  information-rich 3d model repository,'' {\em arXiv preprint
  arXiv:1512.03012}, 2015.

\bibitem{chen2021equivariant}
H.~Chen, S.~Liu, W.~Chen, H.~Li, and R.~Hill, ``Equivariant point network for
  3d point cloud analysis,'' in {\em Proceedings of the IEEE/CVF conference on
  computer vision and pattern recognition}, pp.~14514--14523, 2021.

\bibitem{li2021leveraging}
X.~Li, Y.~Weng, L.~Yi, L.~J. Guibas, A.~Abbott, S.~Song, and H.~Wang,
  ``Leveraging se (3) equivariance for self-supervised category-level object
  pose estimation from point clouds,'' {\em Advances in neural information
  processing systems}, vol.~34, pp.~15370--15381, 2021.

\bibitem{wang2019normalized}
H.~Wang, S.~Sridhar, J.~Huang, J.~Valentin, S.~Song, and L.~J. Guibas,
  ``Normalized object coordinate space for category-level 6d object pose and
  size estimation,'' in {\em Proceedings of the IEEE/CVF Conference on Computer
  Vision and Pattern Recognition}, pp.~2642--2651, 2019.

\bibitem{wu20153d}
Z.~Wu, S.~Song, A.~Khosla, F.~Yu, L.~Zhang, X.~Tang, and J.~Xiao, ``3d
  shapenets: A deep representation for volumetric shapes,'' in {\em Proceedings
  of the IEEE conference on computer vision and pattern recognition},
  pp.~1912--1920, 2015.

\bibitem{rad2017bb8}
M.~Rad and V.~Lepetit, ``Bb8: A scalable, accurate, robust to partial occlusion
  method for predicting the 3d poses of challenging objects without using
  depth,'' in {\em Proceedings of the IEEE international conference on computer
  vision}, pp.~3828--3836, 2017.

\bibitem{kehl2017ssd}
W.~Kehl, F.~Manhardt, F.~Tombari, S.~Ilic, and N.~Navab, ``Ssd-6d: Making
  rgb-based 3d detection and 6d pose estimation great again,'' in {\em
  Proceedings of the IEEE international conference on computer vision},
  pp.~1521--1529, 2017.

\bibitem{xiang2017posecnn}
Y.~Xiang, T.~Schmidt, V.~Narayanan, and D.~Fox, ``Posecnn: A convolutional
  neural network for 6d object pose estimation in cluttered scenes,'' {\em
  arXiv preprint arXiv:1711.00199}, 2017.

\bibitem{park2020latentfusion}
K.~Park, A.~Mousavian, Y.~Xiang, and D.~Fox, ``Latentfusion: End-to-end
  differentiable reconstruction and rendering for unseen object pose
  estimation,'' in {\em Proceedings of the IEEE/CVF conference on computer
  vision and pattern recognition}, pp.~10710--10719, 2020.

\bibitem{wang20206}
C.~Wang, R.~Mart{\'\i}n-Mart{\'\i}n, D.~Xu, J.~Lv, C.~Lu, L.~Fei-Fei,
  S.~Savarese, and Y.~Zhu, ``6-pack: Category-level 6d pose tracker with
  anchor-based keypoints,'' in {\em 2020 IEEE International Conference on
  Robotics and Automation (ICRA)}, pp.~10059--10066, IEEE, 2020.

\bibitem{he2020pvn3d}
Y.~He, W.~Sun, H.~Huang, J.~Liu, H.~Fan, and J.~Sun, ``Pvn3d: A deep point-wise
  3d keypoints voting network for 6dof pose estimation,'' in {\em Proceedings
  of the IEEE/CVF conference on computer vision and pattern recognition},
  pp.~11632--11641, 2020.

\bibitem{fischler1981random}
M.~A. Fischler and R.~C. Bolles, ``Random sample consensus: a paradigm for
  model fitting with applications to image analysis and automated
  cartography,'' {\em Communications of the ACM}, vol.~24, no.~6, pp.~381--395,
  1981.

\bibitem{hinterstoisser2011multimodal}
S.~Hinterstoisser, S.~Holzer, C.~Cagniart, S.~Ilic, K.~Konolige, N.~Navab, and
  V.~Lepetit, ``Multimodal templates for real-time detection of texture-less
  objects in heavily cluttered scenes,'' in {\em 2011 international conference
  on computer vision}, pp.~858--865, IEEE, 2011.

\bibitem{calli2015ycb}
B.~Calli, A.~Singh, A.~Walsman, S.~Srinivasa, P.~Abbeel, and A.~M. Dollar,
  ``The ycb object and model set: Towards common benchmarks for manipulation
  research,'' in {\em 2015 international conference on advanced robotics
  (ICAR)}, pp.~510--517, IEEE, 2015.

\bibitem{poulenard2019effective}
A.~Poulenard, M.-J. Rakotosaona, Y.~Ponty, and M.~Ovsjanikov, ``Effective
  rotation-invariant point cnn with spherical harmonics kernels,'' in {\em 2019
  International Conference on 3D Vision (3DV)}, pp.~47--56, IEEE, 2019.

\bibitem{fuchs2020se3transformers}
F.~B. Fuchs, D.~E. Worrall, V.~Fischer, and M.~Welling, ``Se(3)-transformers:
  3d roto-translation equivariant attention networks,'' 2020.

\bibitem{ma2022rethinking}
X.~Ma, C.~Qin, H.~You, H.~Ran, and Y.~Fu, ``Rethinking network design and local
  geometry in point cloud: A simple residual mlp framework,'' {\em arXiv
  preprint arXiv:2202.07123}, 2022.

\bibitem{huynh2009metrics}
D.~Q. Huynh, ``Metrics for 3d rotations: Comparison and analysis,'' {\em
  Journal of Mathematical Imaging and Vision}, vol.~35, pp.~155--164, 2009.

\bibitem{butt1998optimum}
M.~A. Butt and P.~Maragos, ``Optimum design of chamfer distance transforms,''
  {\em IEEE Transactions on Image Processing}, vol.~7, no.~10, pp.~1477--1484,
  1998.

\bibitem{qi2017pointnet}
C.~R. Qi, H.~Su, K.~Mo, and L.~J. Guibas, ``Pointnet: Deep learning on point
  sets for 3d classification and segmentation,'' in {\em Proceedings of the
  IEEE conference on computer vision and pattern recognition}, pp.~652--660,
  2017.

\bibitem{thomas2019kpconv}
H.~Thomas, C.~R. Qi, J.-E. Deschaud, B.~Marcotegui, F.~Goulette, and L.~J.
  Guibas, ``Kpconv: Flexible and deformable convolution for point clouds,'' in
  {\em Proceedings of the IEEE/CVF international conference on computer
  vision}, pp.~6411--6420, 2019.

\end{thebibliography}

\end{document}